# EnchantedClothes: Visual and Tactile Feedback with an Abdomen-Attached Robot through Clothes

Takumi Yamamoto[1], Rin Yoshimura[1], and Yuta Sugiura[1]

[1] *Keio University, Yokohama, Japan*

(Email: imuka06x17@keio.jp)

**Abstract ---** Wearable robots are designed to be worn on the human body. Taking advantage of their physical form, various applications for wearable robots are being considered. This study proposes a wearable robot worn on the abdomen and a new interaction with it. Our robot enables a variety of applications related to communication between the wearer and surrounding humans through visual and tactile feedback. The contributions of this research will be (1) the proposal of a novel wearable robot worn on the abdomen and (2) a new interaction with it.

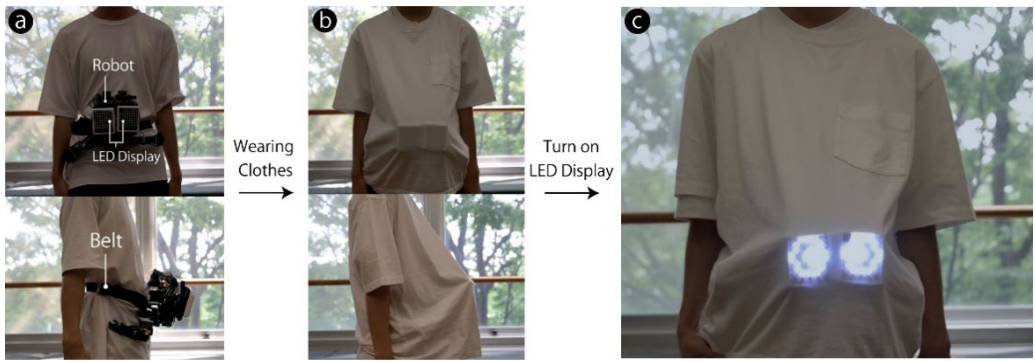

Fig.1 Examples of our proposed robot use. (a) First, the user wears the robot on the abdomen using a belt. (b) Next, the user puts on clothes over the robot. (c) LED display equipped with a robot is shined, or moves under the clothes worn.

## 1 INTRODUCTION

A wearable robot is a form of robot that is worn on the human body and is being considered for a variety of applications. For instance, these robots can augment human efforts in real-world physical tasks. In this use case, the wearer can operate the robot from a remote location or use various parts of their body to control it. Another example is when the robot is used as an agent to communicate with the wearer and people around them through physical presence and tactile sensation. In the future, wearable robots are expected to act as agents to provide work support and communication. The location of wearable robots can be diverse. Examples include those worn on the shoulder for navigation [1], attached to the arm [2], and enabling full-body movement [3, 4]. This study focuses on the potential of wearable robots attached to the abdomen. There is an animation called "A Gutsy Frog" [5] that was broadcast on television in Japan from 1981 to 1982. In this animation, a living frog attaches to the protagonist's clothes and lives inside them, solving the challenges they face together. Strongly inspired by this concept, this research proposes a novel wearable robot that presents visual and tactile feedback through clothes.

We propose a wearable robot attached to the abdomen and a new interaction using the robot and clothes. Users attach the robot to their abdomen and wear clothes over the attached robot. The robot is equipped with an LED display, which provides visual feedback to the user through the clothes. Furthermore, the robot provides visual and tactile effects by its physical movements through the clothes. Various applications relating to communication between the wearer and the people around them can be expected due to the visual feedback created by this robot. Additionally, by pulling on the clothes, the robot can indicate direction to the user via tactile sensation, thereby enabling route navigation. The contributions of this study are (1) the proposal of a novel wearable robot worn on the abdomen, which provides tactile and visual feedback, and the implementation of a prototype, and (2) the proposal of interactions between users and the robot through the clothes, such as communication and navigation.

## 2 PROPOSED METHODS

### 2.1 Design

In this study, we propose a wearable robot that is attached to the abdomen, enabling visual and tactile feedback through clothes. Our robot's design is inspired by the character "Pyonkichi" from the popular manga, anime, film, and drama series "The Gutsy Frog" [5] in Japan. Pyonkichi is a character printed on clothes, but there are scenes where he interacts emotionally with people, conversing with them and pulling on their clothes. Similar to humans, Pyonkichi grows alongside others through experiences of conflict and mutual assistance, becoming a companion-like presence for the wearer.

Inspired by Pyonkichi, we have designed a wearable robot that wraps around the abdomen. People alleviate negative emotions by touching soft things [6] and become attached to devices when they wear them [7]. We believed that presenting information through soft clothing worn daily can appeal to people's hearts and minds, and attaching the robot to the abdomen, which has a large surface area, enables a stronger presentation of emotions to non-wearers. In addition, other advantages of this design are that, when inactive, the robot seamlessly integrates into the clothes without causing any inconvenience. Additionally, by changing the clothing, the overall impression of the robot can be altered, thereby expanding the design possibilities.

### 2.2 Prototype Implementation

For interaction with users through clothes, users should attach an actuator to the abdomen. In our prototype, an existing biped robot (KHR-3HV22 Ver.3, Kondo Kagaku) was used as the actuator. KHR-3HV22 consists of 22 servo motors and can move freely on 22 axes. The motion was designed using Heart To Heart (HTH4) software, which is compatible with KHR-3HV22. HTH4 allows for no-code and intuitive motion creation by combining and wiring panels. A teaching function exists, allowing actual robot movements to be reflected in the panel. After designing the motion, we can upload the motion data to the controller board (RCB-4HV).

To provide visual feedback to the users, two LED matrix displays (Fusing Kitronik ZIP Tile from Kitronik) are attached to the tip of each of the robot's legs, allowing the LED light to penetrate the clothes and be seen by the user, similar to [8]. The circuit diagram is shown in Fig 2. The robot (KHR-3HV22) is controlled by the RCB-4HV, and the LED display is controlled by the Microbit. These two microcontrollers are wired to a PC (ASUS ZenBook S (UX393EA-HK001TS)) for serial communication with a Python program. By entering commands into the Python program, the LED displays and the robot can be controlled.

The flow of robot attachment is illustrated in Fig 1. First, as shown in Fig 1 (a), the user fixed the robot to the abdomen with its head facing downward. We used a belt to fix the robot on the abdomen. Then, as shown in Fig 1 (b), the wearer puts on the clothes. Finally, as shown in Fig 1 (c), the user completes the setup by adjusting the clothes to ensure the robot is properly covered and positioned. It is possible to provide visual feedback using LEDs through the worn clothes. Furthermore, tactile feedback can also be realized by moving the clothes.

### 2.3 Application Scenarios

Our robot can act as an agent to navigate the wearers. To achieve this, the robot needs to communicate with others, interact with the wearer, and provide guidance. The following describes these functions. By wearing this robot, the wearer can enhance their emotional expression. As shown in Fig. 3, we designed facial expressions using an LED display and motions using actuators corresponding to delighted (Fig. 3 (b)), sad (Fig. 3 (c)), and happy (Fig. 3 (d)). For details on the motion, please refer to the supplementary video. Expanding emotional expression using this robot can be considered beneficial for promoting smooth communication.

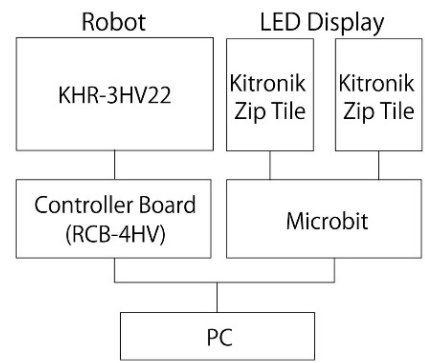

Fig 2. Circuit diagram

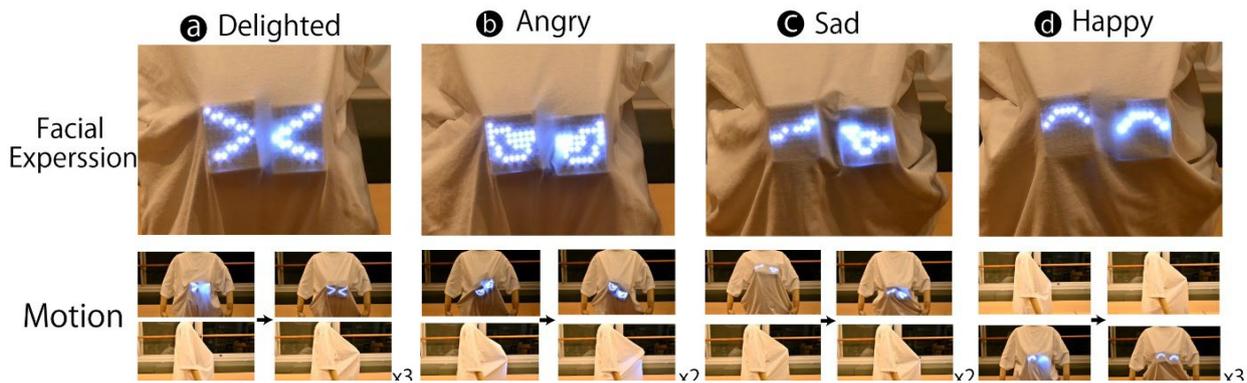

Fig.3  Four types of emotions ((a) Delighted, (b) Angry, (c) Sad, (d) Happy) are presented through the robot's motions and LED display lighting. This enhanced the augmentation of wearer's Emotions.

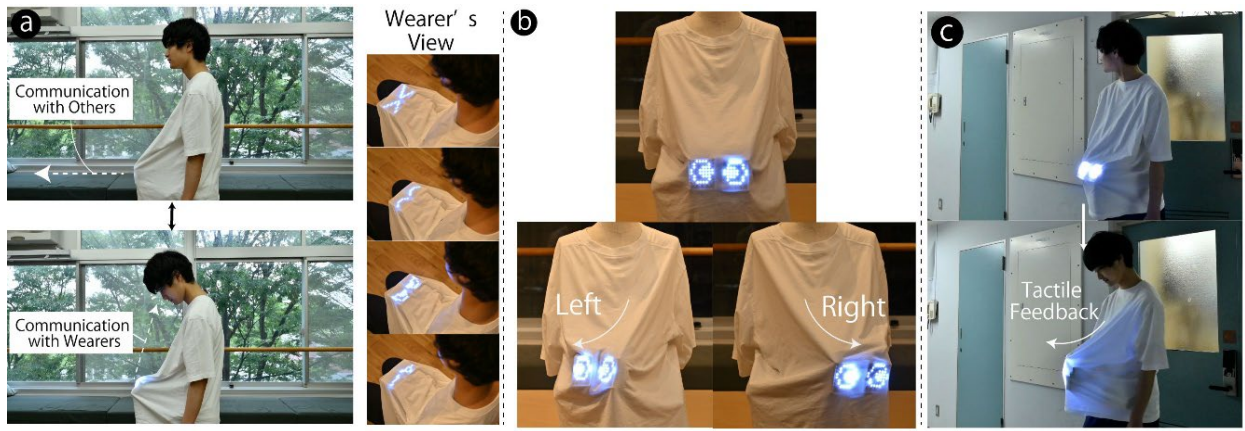

Fig.4 (a) This robot can communicate with not only others but also wearers. By changing the angle of its leg, it can switch the target of expressing its emotions. (b) This robot changes the horizontal angle of the robot's legs. (c) By changing the angle of the robot's legs, the robot can navigate the wearer through tactile feedback.

In addition, the wearer can also interact with the robot itself. Fig. 4 (a) shows how this is done. The robot can change whether it communicates with the wearer or non-wearer by changing the angle of its legs. By tilting the angle of the legs upward, the wearer can look at the LED display and understand the facial expressions on it, thus communicating the robot's emotions to the wearer. The wearer of a robot receives a positive impression of the robot by wearing it [7]. In this study, we believe that this function will make the wearer feel more attached to the robot. This robot can change the horizontal angle of its legs as shown in Fig. 4 (b), thereby changing the left and right directions in which it pulls the clothes. This allows for navigation, such as guiding the wearer along the way. The robot can guide the user in the direction the user wants to go in the real world, as shown in Fig. 4 (c).

## 3   PRELIMINARY USER FEEDBACK

We gathered user feedback to explore the concept and identify areas for improvement in our early prototype. We recruited five students (3 males, mean age 23.8 years (SD = 2.17)). The experimental participants first received an explanation of the device and then wore it. After wearing the device, the robot moved as shown in Fig. 3. After that, we conducted a semi-structured interview with two questions: (1) How did you feel about the size and weight of the device? (2) Do you have any opinions on the body part where the device is worn?

As a result, many participants expressed concerns about the size of the device. Regarding the weight, most participants found it to be within an acceptable range, although some expressed concerns. For example, P1 said, "I didn't feel it was too heavy. However, the size felt large." In addition, two participants also mentioned how the device was worn and secured. Although the device is attached to the body using a belt, the weight and balance of the device may have been unbalanced, making the fixation unstable. For example, P2 said, "I would like some support at the fixed position, specifically where it corresponds to the hip joint of a bipedal walking robot." Regarding the position of the device, three of the five participants had a positive or non-negative impression of wearing the device on the abdomen. One expressed concern about use by women, although two female participants did not express concern. P4 said, "Women may not like to wear it.".

## 4 DISCUSSION, LIMITATIONS AND FUTURE WORK

In this study, we implemented the prototype of the robot using KHR-3HV2. The size of this robot is 401.05 mm × 194.4 mm × 129 mm, and participants of the preliminary user study expressed concerns about its size. Because of its large size, the portability of the wearable robot is compromised, and it may get in the way when performing other tasks. In the future, we will redesign the prototype to make it smaller and lighter to improve portability. In particular, we are considering a device that can be folded, as described in [9]. Furthermore, the resolution of the LED display used in this study is limited to $8 \times 8$. Higher resolution may enable a variety of presentations, and we will aim to achieve higher resolution in the future.

In this research, the robot is used to move clothes, which may result in the clothes being stretched. Also, depending on the color and material of the clothes, the LED light may not be projected properly to the clothes. We will investigate the effects of continuous use on clothes and identify materials that can better transmit LED light.

Currently, the robot is capable of providing tactile and visual feedback to users. In the future, it will be possible to realize further interaction by enabling not only these modalities but also different types of presentation (e.g., voice). In addition, although the robot currently provide the feedback to the user, we believe that further interaction in the real world can be created by changing the content of the presentations based on the measurement of the user's state and input from the user. In the future, we will consider a method in which the pressure sensor is installed on the robot so that the user can provide input to the robot.

## 5 CONCLUSION

In this study, inspired by the character "Pyonkichi" from "The Gutsy Frog" [25], we propose a novel abdomen-attached robot that provides visual and tactile feedback to users through clothes. We propose various applications relating to communication with this robot. In addition, this robot can guide the user in the direction via tactile sensation, thereby enabling route navigation. This robot allows users to communicate in a new way through clothing and enhance real-world tasks. However, participants in the preliminary user study raised concerns about the design of the robot such as its size, position on the body, discomfort for women. In the future, we will refine the robot's design to enhance comfort, portability, and usability.

ACKNOWLEDGEMENT

Part of this work was supported by JST SPRING, Grant Number JPMJSP2123. This work was supported by JSPS KAKENHI Grant Numbers JP23H01046, JP24KJ1957.

REFERENCES

[1] T. Kashiwabara, H. Osawa, K. Shinozawa, and M. Imai, "TEROOS: A Wearable Avatar to Enhance Joint Activities," in *Proc. SIGCHI Conf. Human Factors in Computing Systems* (CHI '12), Austin, TX, USA, 2012, pp. 2001-2004.

[2] S. Yoshida, T. Sasaki, Z. Kashino, and M. Inami, "TOMURA: A Mountable Hand-Shaped Interface for Versatile Interactions," in Proc. Augmented Humans Int. Conf. (AHs '23), Glasgow, United Kingdom, 2023, pp. 243-254.

[3] A. Dementyev, J. Hernandez, S. Follmer, I. Choi, and J. Paradiso, "SkinBot: A Wearable Skin Climbing Robot," in Adjunct Proc. 30th Annu. ACM Symp. User Interface Software and Technology (UIST '17 Adjunct), Québec City, QC, Canada, 2017, pp. 5-6.

[4] A. Dementyev, H.-L. (C.) Kao, I. Choi, D. Ajilo, M. Xu, J. A. Paradiso, C. Schmandt, and S. Follmer, "Rovables: Miniature On-Body Robots as Mobile Wearables," in Proc. 29th Annu. Symp. User Interface Software and Technology (UIST '16), Tokyo, Japan, 2016, pp. 111-120.

[5] Y. Yoshizawa, "The Gutsy Frog", Syueisya, 1970-1976.

[6] Y. Yim, Y. Noguchi, and F. Tanaka, "A wearable soft robot that can alleviate the pain and fear of the wearer," Scientific Reports, vol. 12, no. 1, p. 17003, 2022.

[7] N. Matsunaga and M. Shiomi, "Does a wearing change perception toward a robot?," in 2021 30th IEEE Int. Conf. Robot & Human Interactive Communication (RO-MAN), 2021, pp. 963-968.

[8] A. A. R. Irudayaraj, R. Agarwal, N. Joshi, A. Gupta, O. 63Abari, and D. Vogel, "PocketView: Through-Fabric Information Displays," in 34th Annu. ACM Symp. User Interface Software and Technology (UIST '21), Virtual Event, USA, 2021, pp. 511-523.

[9] Z. Ding, S. Yoshida, T. Torii, and H. Xie, "XLimb: Wearable Robot Arm with Storable and Extendable Mechanisms," in *12th Augmented Human Int. Conf.* (AH2021), Geneva, Switzerland, 2021, Article 8, 4 pages.